\documentclass[10pt,twocolumn,letterpaper]{article}

\usepackage[final]{cvpr} 

\usepackage{graphicx, amsmath, amssymb, caption, subcaption, multirow, overpic}
\usepackage[table,dvipsnames]{xcolor}
\usepackage[british, english, american]{babel}
\usepackage{algorithm}
\usepackage{listings}
\usepackage{textpos}

\usepackage{etoolbox}
\makeatletter
\AfterEndEnvironment{algorithm}{\let\@algcomment\relax}
\AtEndEnvironment{algorithm}{\kern2pt\hrule\relax\vskip3pt\@algcomment}
\let\@algcomment\relax
\newcommand\algcomment[1]{\def\@algcomment{\footnotesize#1}}
\renewcommand\fs@ruled{\def\@fs@cfont{\bfseries}\let\@fs@capt\floatc@ruled
  \def\@fs@pre{\hrule height.8pt depth0pt \kern2pt}%
  \def\@fs@post{}%
  \def\@fs@mid{\kern2pt\hrule\kern2pt}%
  \let\@fs@iftopcapt\iftrue}
\makeatother

\usepackage[pagebackref=false, breaklinks=true, letterpaper=true, colorlinks,
            citecolor=citecolor, linkcolor=linkcolor, bookmarks=false]{hyperref}
\definecolor{citecolor}{HTML}{0071BC}
\definecolor{linkcolor}{HTML}{ED1C24}

\def\cvprPaperID{122}
\def\confName{CVPR}
\def\confYear{2022}

\newlength\savewidth\newcommand\shline{\noalign{\global\savewidth\arrayrulewidth
  \global\arrayrulewidth 1pt}\hline\noalign{\global\arrayrulewidth\savewidth}}
\newcommand{\tablestyle}[2]{\setlength{\tabcolsep}{#1}\renewcommand{\arraystretch}{#2}\centering\footnotesize}
\renewcommand{\paragraph}[1]{\vspace{1.25mm}\noindent\textbf{#1}}

\newcolumntype{x}[1]{>{\centering\arraybackslash}p{#1pt}}
\newcolumntype{y}[1]{>{\raggedright\arraybackslash}p{#1pt}}
\newcolumntype{z}[1]{>{\raggedleft\arraybackslash}p{#1pt}}

\newcommand{\app}{\raise.17ex\hbox{$\scriptstyle\sim$}}

\definecolor{deemph}{gray}{0.6}

\definecolor{baselinecolor}{gray}{.92}

\newcommand{\authorskip}{\hspace{2.5mm}}

\begin{document}
\title{\vspace{-1mm}\Large SLIP: Self-supervision meets Language-Image Pre-training \vspace{-3mm}}
\author{
 Norman Mu$^{1}$ \authorskip Alexander Kirillov$^{2}$ \authorskip David Wagner$^{1}$ \authorskip Saining Xie$^{2}$ \\[2mm]
 $^1$UC Berkeley,\quad $^2$Facebook AI Research (FAIR)}

\maketitle

\begin{abstract}
Recent work has shown that self-supervised pre-training leads to improvements over supervised learning on challenging visual recognition tasks.
CLIP, an exciting new approach to learning with language supervision, demonstrates promising performance on a wide variety of benchmarks.
In this work, we explore whether self-supervised learning can aid in the use of language supervision for visual representation learning. 
We introduce SLIP, a multi-task learning framework for combining self-supervised learning and CLIP pre-training.
After pre-training with Vision Transformers, we thoroughly evaluate representation quality and compare performance to both CLIP and self-supervised learning under three distinct settings: zero-shot transfer, linear classification, and end-to-end finetuning. Across ImageNet and a battery of additional datasets, we find that SLIP improves accuracy by a large margin.
We validate our results further with experiments on different model sizes, training schedules, and pre-training datasets.
Our findings show that SLIP enjoys the best of both worlds: better performance than self-supervision (+8.1\% linear accuracy) and language supervision (+5.2\% zero-shot accuracy).
\end{abstract}\hspace{-1.8em}

\begin{textblock*}{.8\textwidth}[.5,0](0.5\textwidth, -.625\textwidth)
\centering
{\small Code: \url{https://github.com/facebookresearch/SLIP}}
\end{textblock*}

\section{Introduction}
\label{sec:intro}
Much of the recent progress in deep learning has been driven by the paradigm of pre-training powerful, general-purpose representations that transfer well to a variety of specific applications.
Within computer vision, supervised learning on image classification and self-supervised learning on unlabeled images comprise the two primary approaches to representation learning.
After AlexNet \cite{Krizhevsky2012ImageNetCW}, researchers soon realized that supervised pre-training yields a generic visual backbone which can be repurposed for many different tasks \cite{Girshick2014RichFH}.
Today, most state-of-the-art results still depend on supervised pre-training, and scaling to massive amounts of data, such as Google's proprietary JFT dataset, remains one of the most reliable methods for improving downstream performance.
Self-supervised learning, a form of unsupervised learning, found tremendous success first in the domain of language \cite{Devlin2019BERTPO,Radford2018ImprovingLU}, but has also made significant recent progress in vision.
A major motivation for studying self-supervised learning has been a desire to supersede supervised pre-training and its reliance on labor-intensive human annotation.
Indeed, self-supervised pre-training has outperformed supervised learning for some time now on small datasets, but only recently with the development of contrastive methods \cite{Chen2020ASF,He2020MomentumCF} has it begun to improve performance on larger datasets such as ImageNet.

\begin{figure}[t]
\centering
\hspace{-.5em}
\includegraphics[width=1.0\linewidth]{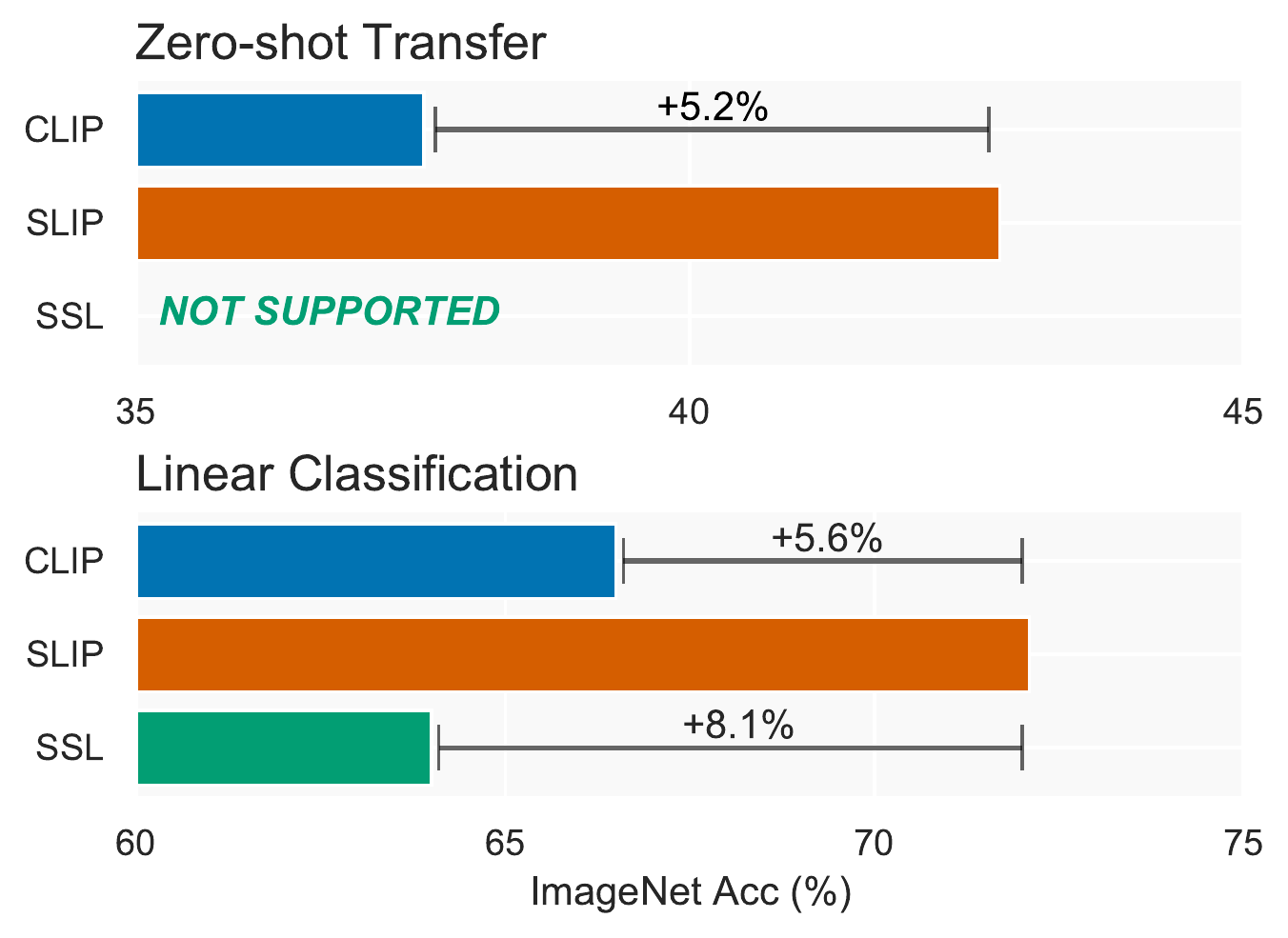}
\vspace{-.5em}
\caption{\textbf{SLIP pre-training on YFCC15M.} Combining image-only self-superivsion and image-text supervision simultaneously improves zero-shot transfer and linear classification on ImageNet.
}
\label{fig:teaser}
\vspace{-1.5em}
\end{figure}

Both supervised and self-supervised pre-training today rely heavily on ImageNet (\ie ImageNet-1K)~\cite{Russakovsky2015ImageNetLS}, a highly curated dataset with particular idiosyncrasies and biases \cite{Torralba2011UnbiasedLA}.
The YFCC100M dataset \cite{Thomee2016YFCC100MTN} was released in 2015 and remains the largest publicly-accessible collection of images.
To date, the field of representation learning has found much less use for this dataset.
On the other hand, the full ImageNet dataset of 14M images (\ie ImageNet-22K) has become very popular for its role in training Vision Transformer models which require a larger amount of data than ImageNet-1K~\cite{Dosovitskiy2021AnII,Bao2021BEiTBP}.
Why are uncurated datasets not more common in the study of representation learning?
There are a few possible reasons.
Most immediately, uncurated datasets also lack labels and so long as supervised pre-training remains the simpler and more accessible option for most researchers, datasets like YFCC100M are a non-starter.
As we confirm again later in this work, the standard self-supervised evaluation task of ImageNet classification from frozen features heavily biases results against models not also pre-trained on ImageNet \cite{Caron2018DeepCF}.
Finally, while progress on ImageNet has been encouraging, there has not been strong evidence that current self-supervised methods scale well to larger uncurated datasets~\cite{tian2021divide}.

Recently, CLIP \cite{Radford2021LearningTV} introduced an exciting new approach to representation learning.
It re-examines language supervision for learning visual representations, and catapults it into contention with label supervision and self-supervision.
CLIP requires only images and free-form text captions, thus revitalizing the use of YFCC100M in representation learning.
In addition to no longer requiring label annotations, CLIP accuracy also scales well to large datasets and models.
The best results for CLIP are achieved with big models on a curated dataset of 400M images and captions, though promising results are also shown on a subset of YFCC100M.
CLIP also enables many exciting new applications with its flexible language-guided capabilities.

\begin{figure}[t]
\centering
\hspace{-.5em}
\includegraphics[width=0.9\linewidth]{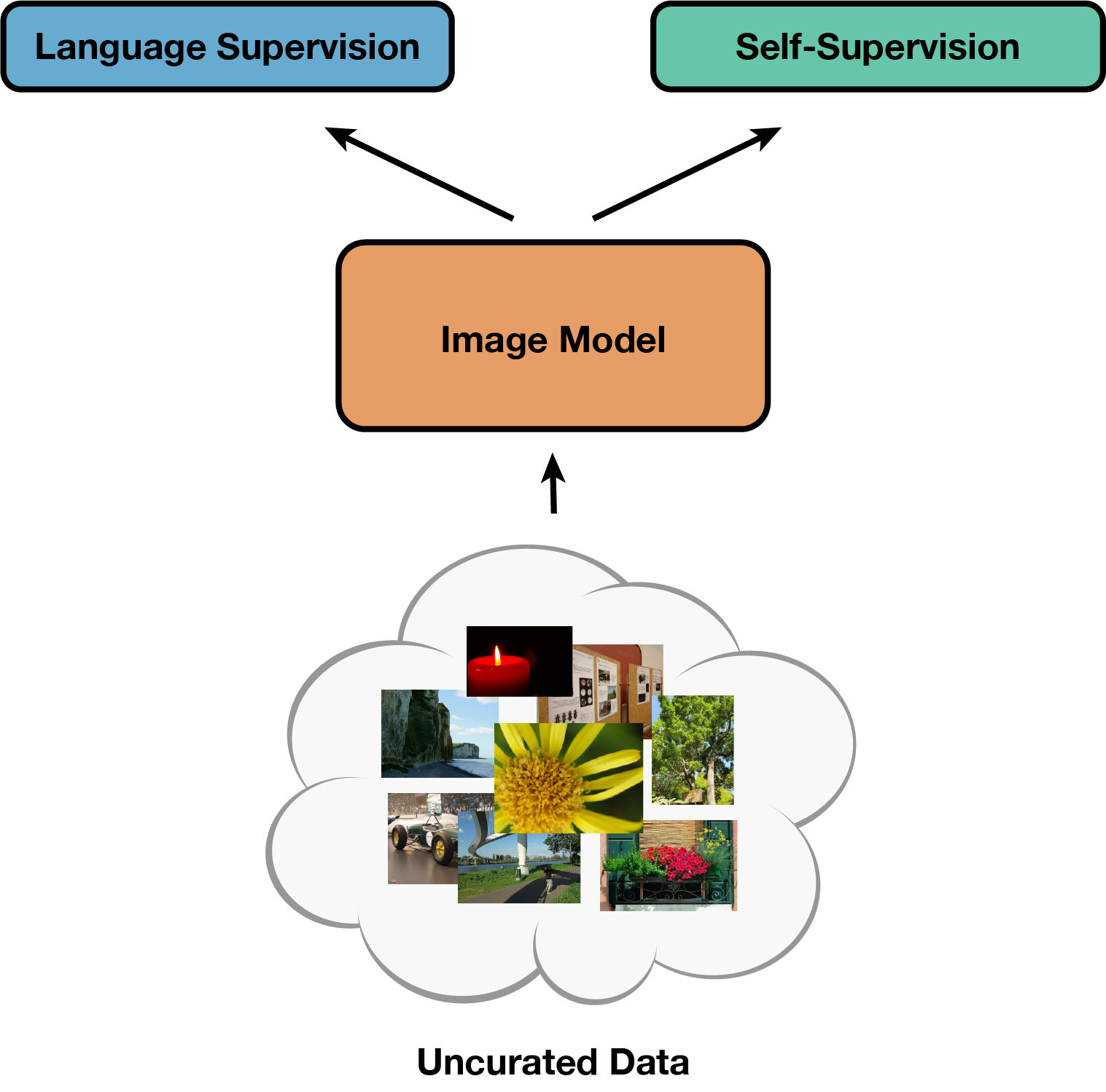}
\vspace{-.5em}
\caption{Illustration of SLIP, our multi-task framework. An image model has access to and can be trained with both language supervision from captions and self-supervision on images.}
\label{fig:cartoon}
\end{figure}

In this work, we explore whether the momentum of self-supervised learning on images carries into the setting of language supervision.
In particular, we investigate whether language supervision in the form of CLIP also benefits from image self-supervision.
We note that it is not immediately clear that these two training objectives should be stronger together.
The two objectives each require the model to encode qualitatively different and conflicting information about the image, leading to interference.

In order to explore these questions, we introduce SLIP (Self-supervision meets Language-Image Pre-training), a multi-task framework combining language supervision and self-supervision.
We pre-train various SLIP models on a subset of YFCC100M, and thoroughly evaluate representation quality under three distinct settings: zero-shot transfer, linear classification, and end-to-end finetuning. 
We evaluate downstream performance on ImageNet, in addition to a battery of 25 other classification benchmarks.
Additionally, we further validate our findings with experiments on different model sizes, training schedules, and pre-training datasets.
Our findings conclusively show that SLIP improves performance across most evaluations by a significant margin, an encouraging signal for the general utility of self-supervision in the context of language supervision.
Additionally, we analyze various components of our method in further detail such as the choices of pre-training dataset and data processing method.
We conclude with a discussion of our evaluations as well as the ethical and practical limitations of this class of methods.

\section{Related Work}

\paragraph{Language supervision.}
Early work explored learning visual representations from image captions, even before the advent of deep learning \cite{Quattoni2007LearningVR}.
DeViSE \cite{Frome2013DeViSEAD} jointly embeds images and textual class labels within a shared semantic space, allowing the model to recognize classes that were not explicitly trained for.
Initial attempts at leveraging the YFCC dataset for representation learning included predicting the bag-of-words representation \cite{Joulin2016LearningVF} or n-gram occurrence \cite{Li2017LearningVN} from images.
ICMLM \cite{Sariyildiz2020LearningVR} and VirTex \cite{Desai2021VirTexLV} showed that language supervision on COCO Captions produced useful visual representations.
Prior to CLIP, Multimodal Contrastive Training \cite{Yuan2021MultimodalCT} adds contrastive image-image and language-image losses to VirTex which further improve performance.
CLIP \cite{Radford2021LearningTV} quickly garnered significant attention for its simplicity, scale, and strong results.
Developed concurrently, ALIGN \cite{Jia2021ScalingUV}, uses a larger but noisier uncurated dataset and shows similar results.

\paragraph{Self-supervised learning.}
Earlier self-supervised learning methods have shown subpar scaling with dataset size~\cite{Goyal2019ScalingAB}. Contrastive learning methods ushered in rapid progress \cite{Oord2018RepresentationLW,Wu2018UnsupervisedFL,He2020MomentumCF,Chen2020ASF} due to their simplicity and effectiveness. Recent methods for self-supervised learning also propose a variety of alternatives to the contrastive objective such as self-distillation \cite{Grill2020BootstrapYO,Caron2021EmergingPI}, or input reconstruction \cite{Bao2021BEiTBP,He2021MaskedAA}.

\paragraph{Multi-modal multi-task learning.}
MURAL \cite{Jain2021MURALMM} extends ALIGN to the multi-lingual setting and introduces a cross-lingual objective to improve multi-lingual image and text retrieval.
Concurrently to this work, DeCLIP \cite{Li2021SupervisionEE} adds several additional training objectives and more data collected in-house to CLIP in order to improve data efficiency.

\section{SLIP Framework}
\label{sec:slip}

We introduce SLIP, a framework for combining language supervision and image self-supervision to learn visual representations without category labels.
During pre-training, separate views of each input image are constructed for the language supervision and image self-supervision branches, then fed through a shared image encoder.
Through the course of training, the image encoder learns to represent visual input in a semantically meaningful manner.
We can then measure the quality of these learned representations by evaluating their utility in downstream tasks.

\subsection{Contrastive Language-Image Pre-training}

Radford et al. \cite{Radford2021LearningTV} demonstrated the ability of contrastive learning (CLIP) on corresponding images and captions to learn powerful visual representations.
CLIP first embeds images and text with separate modality-specific models.
These vectors are then projected into a shared embedding space and normalized.
The InfoNCE loss is computed using these final embeddings, with corresponding images and captions as positive pairs and all non-matching images and captions as negative pairs.

Non-contrastive alternatives for language supervision include predicting the bag-of-words representation of the caption \cite{Joulin2016LearningVF} or the original caption \cite{Sariyildiz2020LearningVR,Desai2021VirTexLV} from the image.
However, these methods appear to yield weaker results than CLIP.
The contrastive objective also enables image classification without re-training dataset-specific classification layers (zero-shot transfer).

\subsection{Image Self-Supervision}
\label{sec:slip_ssl}

View-based self-supervised learning, in which models are trained to represent views or augmentations of the same image similarly, has yielded strong results across a variety of different formulations.
In this work we primarily use an adaptation of SimCLR~\cite{Chen2020ASF,Chen2020BigSM}, a representative example of these methods, as the self-supervised objective in SLIP.
However, other frameworks can be swapped in quite easily, and we explore this in Section \ref{sec:analysis}.
We focus on the Vision Transformer \cite{Dosovitskiy2021AnII} architecture for its simplicity and good performance. We follow MoCo v3 \cite{Chen2021AnES} on the hyperparameter settings for training self-supervised Vision Transformers, which will be described later in Section~\ref{sec:implementation}.

\subsection{Our Method}
We outline SLIP with SimCLR for self-supervision (i.e. SLIP-SimCLR) in Algorithm \ref{alg:code}.
During each forward pass in SLIP, all images are fed through the same encoder.
The CLIP and SSL objectives are computed on the relevant embeddings and then summed together into a single scalar loss.
The two objectives can be balanced differently by rescaling the SSL objective.
We find that a scale of 1.0 for the self-supervised objective, i.e. no re-scaling, works well for SimCLR.
Unless otherwise noted, we refer to SLIP-SimCLR simply as SLIP.

SLIP increases the number of images processed which results in approximately 3$\times$ more activations.
This expands the model's memory footprint and slows down the forward pass during training.
See Section \ref{sec:discussion} for further discussion. 

\section{Improved Training Procedure}

The authors of CLIP focus primarily on training with a large private dataset of 400M image-text pairs, where the large scale of data lessens the need for regularization and data augmentation.
While re-implementing CLIP, we found some simple adjustments primarily to data augmentation which significantly improved performance when pre-trained on YFCC15M.
Our improved training procedure achieves 34.6\% zero-shot transfer to ImageNet with a modified\footnote{The initial 7$\times$7 conv is replaced by three 3$\times$3 convs; global average pooling is replaced by a self-attention pooling layer with 14M parameters.} ResNet-50, exceeding the original result of 31.3\%.
Another re-implementation achieves 32.7\% accuracy on ImageNet \cite{ilharco_gabriel_2021_5143773}.
In our experiments we focus primarily on the Vision Transformer model family for their strong scaling behavior~\cite{Dosovitskiy2021AnII}. We train all Vision Transformer models with our improved procedure as well, in order to set strong baselines for comparing our methods.

\subsection{Implementation Details}
\label{sec:implementation}

\paragraph{Datasets.} We focus primarily on a 15M subset of YFCC100M \cite{Thomee2016YFCC100MTN} filtered by Radford et al. \cite{Radford2021LearningTV} consisting of English-only titles and descriptions, which we refer to as YFCC15M.
We also evaluate on Conceptual Captions 3M (CC3M) \cite{Sharma2018ConceptualCA} and Conceptual Captions 12M (CC12M) \cite{Changpinyo2021Conceptual1P}.

\paragraph{Data Augmentation.} During training, we randomly sample a valid caption for each image (i.e. title or description for YFCC15M).
Images for the CLIP branch are randomly resized and cropped to between 50\% and 100\% of the original image, which we refer to as global cropping.
In the self-supervised branch we sample two views with the augmentation from MoCo v3 \cite{Chen2020ASF}.

\paragraph{Architecture.} We use the original ViT-B/16 and ViT-L/16 architectures from the ViT paper~\cite{Dosovitskiy2021AnII} for our image encoders, as well as a ViT-S/16 architecture~\cite{Touvron2021TrainingDI} which is comparable to ResNet-50 in FLOPs and parameters.
For our text encoders, we use the smallest text Transformer model from CLIP which contains 38M parameters and uses byte-pair encoding with a 49K token vocabulary, and maximum context length of 77.

For the CLIP objective, our model projects the image and caption embeddings into a 512-dim space with separate learned linear projections.
In the self-supervised branch, we use the 3-layer MLP projection head with 4096-dim hidden layers to transform the image embeddings into a 256-dim output space.

\paragraph{Training.} We train with a batch size of 4096 and the AdamW optimizer in all our experiments.
Following CLIP, we set the $\beta_2=0.98$ to improve training stability, but we keep $\epsilon=1e-8$. 
We use a weight decay of 0.5 for CLIP and 0.1 for SLIP.
Instead of the custom mixed-precision recipe used in CLIP, we opt for the built-in automatic mixed precision library in PyTorch.

\paragraph{Zero-shot Transfer Evaluation.} We evaluate zero-shot transfer to various classification benchmarks including ImageNet.
We perform prompt ensembling by averaging the caption embeddings for each class across the prompt templates.
This average caption embedding is then used to compute cosine similarity with the image embeddings.
CLIP provides prompt templates and class names for these benchmarks, which we use directly for ease of comparison.

\paragraph{Linear Classification Evaluation.} We use the same setup as MoCo v3 to evaluate linear classification performance.
We use SGD w/ momentum and no weight decay.
On ImageNet, we use a learning rate of 0.01 and on the other downstream datasets we tune the learning rate and report the best result.
We train for 100 epochs and perform standard cropping and flipping augmentations.

\paragraph{End-to-end Finetuning Evaluation.} To finetune our models on ImageNet, we use the training procedure from BeiT \cite{Bao2021BEiTBP}.
This procedure employs significant regularization and data augmentation, as well as layerwise learning rate decay which exponentially decays the learning rate across layers.
We disable relative positional embedding, layer scaling, and average pooling across tokens.
For ViT-B and ViT-S we train for 100 epochs, while on ViT-L we train for 50 epochs.

For finetuning on smaller downstream datasets, we use the simpler DeiT training procedure \cite{Touvron2021TrainingDI}.

\begin{algorithm}[ht]
\caption{SLIP-SimCLR: PyTorch-like Pseudocode}
\label{alg:code}
\algcomment{
\textbf{Notes}: \texttt{@} is the matrix multiplication operator. \texttt{k.T} is \texttt{k}'s transpose.
\texttt{eye} constructs an identity matrix. \texttt{cat} concatenates two matrices. 
}
\definecolor{codeblue}{rgb}{0.25,0.5,0.5}
\definecolor{codekw}{rgb}{0.85, 0.18, 0.50}
\begin{lstlisting}[language=python]
# fi, ft: image, text encoders
# hi, ht: CLIP image, text projectors
# hs: SimCLR projector
# c: SimCLR loss scale
def forward(img, text):
    xi, x1, x2 = crop(img), aug(img), aug(img)
    yt = tokenize(text)

    wi, w1, w2 = fi(xi, x1, x2)
    wt = ft(yt)

    z1, z2 = hs(w1), hs(w2)  # SSL embed: N x C2
    zi, zt = hi(wi), ht(wt)  # CLIP embed: N x C1

    loss = c * simclr(z1, z2) + clip(zi, zt)
    return loss

# s: learnable log logit scale
def clip(zi, zt):
    zi, zt = normalize(zi, zt)
    label = range(N)
    logit = exp(s) * zi @ zt.T

    li = CrossEntropy(logit, label)
    lt = CrossEntropy(logit.T, label)

    loss = (li + lt) / 2
    return loss

# tau: softmax temperature
def simclr(z1, z2):
    z1, z2 = normalize(z1, z2)
    label = range(N)
    mask = eye(N) * 1e9

    logit = z1 @ z2.T
    logit1 = z1 @ z1.T - mask
    logit2 = z2 @ z2.T - mask

    logit1 = cat(logit, logit1)
    logit2 = cat(logit.T, logit2)

    l1 = CrossEntropy(logit1 / tau)
    l2 = CrossEntropy(logit2 / tau)

    loss = (l1 + l2) / 2
    return loss
\end{lstlisting}
\end{algorithm}

\section{Empirical Evaluations}
\label{sec:evaluations}

\subsection{ImageNet Classification}

\begin{figure*}[h]
\centering
\includegraphics[width=0.9\linewidth]{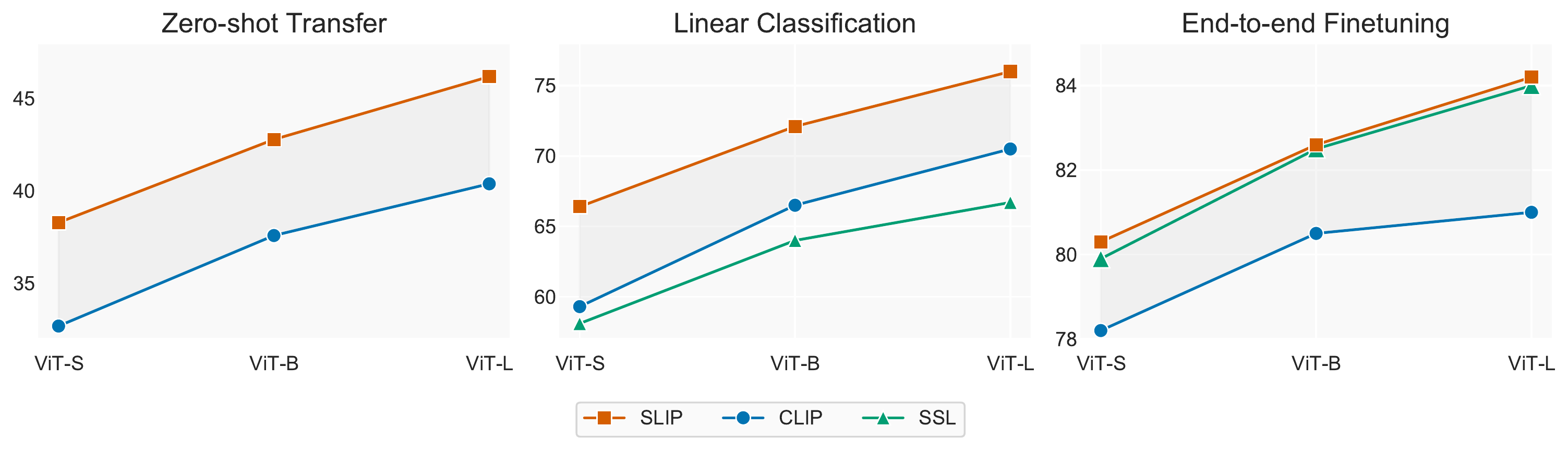}
\vspace{-.5em}
\caption{\textbf{ImageNet results.}
We evaluate the representation quality by testing the performance on ImageNet under different settings: zero-shot transfer using text prompts, linear classification, and end-to-end finetuning.
SLIP improves upon the zero-shot transfer and linear classification performance of CLIP by significant margin across all vision Transformer model sizes.}
\label{fig:main}
\vspace{-.5em}
\end{figure*}

We evaluate performance on ImageNet under three distinct settings: zero-shot transfer, linear classification, and end-to-end finetuning.
The zero-shot transfer task evaluates model performance on classification benchmarks directly after pre-training without updating any of the model weights.
A model trained with contrastive language supervision can be used as an image classifier by simply selecting the class whose caption embedding aligns most closely with the input image.
Linear classification, also called linear probing, is a standard evaluation method used to evaluate unsupervised or self-supervised representations.
A randomly initialized final classification layer is trained while all other model weights are frozen.
Finally, another way of evaluating representation quality is whether a pre-trained model can improve upon the performance of supervised learning when finetuning the model end-to-end.

\begin{table}[h]
\tablestyle{2pt}{1.1}
\setlength{\tabcolsep}{4pt}
\begin{tabular}{l l l l }
Dataset & Method & Linear & Finetuning \\
\shline
ImageNet & SimCLR & 74.5 & 82.8 \\
  & MoCo v3 & 76.6 & 83.1 \\
\hline
YFCC15M & SimCLR & 64.0 \color{Red}(-10.5) & 82.5 \color{Red}(-0.3) \\
  & MoCo v3 & 66.1 \color{Red}(-10.5) & 82.8 \color{Red}(-0.3) \\
\end{tabular}
\vspace{-.5em}
\caption{We train ViT-B/16 with two self-supervised frameworks and find that both linear classification and end-to-end finetuning accuracy on ImageNet suffers when pre-training on YFCC15M instead of ImageNet. Accuracy drop show in \color{Red}(red).
}
\label{tab:shift}
\vspace{-.5em}
\end{table}

One common evaluation setup in the self-supervised learning literature is to train both the model and the linear classifier on ImageNet (i.e. ImageNet-1K), which even without labels is a highly curated and class-balanced dataset.
In Table \ref{tab:shift} we train ViT-B/16 with SimCLR and MoCo v3 on both YFCC15M and ImageNet.
The resulting models are evaluated on ImageNet using linear classification and end-to-end finetuning.
Both SimCLR and MoCo v3 experience a more than 10\% drop in linear classification accuracy when pretrained on YFCC15M instead of ImageNet, a dramatic degradation in performance.
For this reason, the baseline linear results in our experiments are lower than what is typically reported in the self-supervised literature. Similarly, we observe a less severe but consistent degradation for end-to-end finetuning results as well.
We argue that training on uncurated data is a more realistic and informative setting, especially given the original motivations of learning vision from less supervision.

In Table \ref{tab:table1}, we provide evaluation results for CLIP, SimCLR, and SLIP across three sizes of Vision Transformer and on all three ImageNet settings.
All models are trained for 25 epochs on YFCC15M.
We find that language supervision and image self-supervision interact constructively in SLIP, improving upon the performance of both methods alone.

\begin{table}[h]
\tablestyle{2pt}{1.1}
\setlength{\tabcolsep}{6pt}
\begin{tabular}{l l l l l}
Model & Method & 0-shot & Linear & Finetuned \\
\shline
ViT-S/16 & CLIP & \underline{32.7} & \underline{59.3} & 78.2 \\
 & SimCLR & - & 58.1 & \underline{79.9} \\
 & SLIP & \textbf{38.3} \color{Green}(+5.6) & \textbf{66.4} \color{Green}(+7.1) & \textbf{80.3} \color{Green}(+0.4) \\
\hline
ViT-B/16 & CLIP & \underline{37.6} & \underline{66.5} & 80.5 \\
 & SimCLR & - & 64.0 & \underline{82.5} \\
 & SLIP & \textbf{42.8} \color{Green}(+5.2) & \textbf{72.1} \color{Green}(+5.6) & \textbf{82.6} \color{Green}(+0.1) \\
\hline
ViT-L/16 & CLIP & \underline{40.4} & \underline{70.5} & 81.0 \\
 & SimCLR & - & 66.7 & \underline{84.0} \\
 & SLIP & \textbf{46.2} \color{Green}(+4.8) & \textbf{76.0} \color{Green}(+5.5) & \textbf{84.2} \color{Green}(+0.2) \\
\end{tabular}
\vspace{-.5em}
\caption{Full ImageNet results.
SLIP significantly improves performance on ImageNet in the zero-shot transfer, linear classification, and end-to-end finetuning settings.
Improvements over stronger baseline (\underline{underlined}) shown in \color{Green}{green}.
}
\label{tab:table1}
\end{table}

\paragraph{Zero-shot transfer.} Self-supervised models do not support zero-shot transfer evaluation since there is no way to directly map the learned representations onto categorical labels.
SLIP consistently outperforms CLIP by around +5\% on zero-shot transfer across all three model sizes, a very large margin relative to the original number.
The gap between SLIP and CLIP does close slightly between ViT-Small (22M params) and ViT-Large (300M params) from +5.6\% to +4.8\%.
This trend suggests that SLIP would continue to yield benefits over CLIP even for the largest Vision Transformer architectures currently in use.

With ViT-Large SLIP achieves 46.2\% top-1 accuracy, which is far below the performance achieved by smaller models pre-trained on massive curated datasets.
In absolute terms however, this is a very surprising result considering that YFCC15M contains very little data of the specific form seen during zero-shot transfer evaluation (i.e. object-centered images labeled with captions of the form ``a photo of a {class name}''.

\paragraph{Linear Classification.}
\label{sec:linear}
In this setting we also observe the synergy between language supervision and image self-supervision.
CLIP outperforms SimCLR, but by a much smaller margin than SLIP outperforms SimCLR.
We see that SLIP significantly outperforms SimCLR in linear classification accuracy across all three model sizes.
The gap between SLIP and SimCLR is largest with ViT-L at almost +10\%, suggesting that SLIP continues to scale with larger models while SimCLR slightly saturates in performance.

\paragraph{End-to-end Finetuning.}
We see in Table \ref{tab:shift} that finetuning performance is somewhat less affected by pre-training on YFCC15M than linear performance is affected, possibly because the model is allowed to adapt to the target distribution.
Both SimCLR and MoCo v3 experience -0.3\% drops in finetuning accuracy when pre-trained on YFCC15M instead of ImageNet, which is still quite significant for this setting.
We re-iterate that the results in Table \ref{tab:table1} are not directly comparable with methods which are pre-trained on ImageNet-1K.

When finetuning on ImageNet, CLIP is particularly weak: ViT-S and ViT-B performance is below even that of training from a random weight initialization \cite{Touvron2021TrainingDI}.
The performance of CLIP does not scale well with model size either, as CLIP ViT-L performance is only +0.5\% above CLIP ViT-B.
On the other hand, self-supervised learning does quite well in this setting, especially with the larger models.
SimCLR ViT-L enjoys a +3.0\% gain in accuracy over CLIP ViT-L, and SLIP ViT-L does slightly better than SimCLR ViT-L, though by a very marginal amount.
These results suggest that the subpar finetuning performance of CLIP is mostly solved with self-supervision.

\subsection{Model and Compute Scaling}

We also investigate the scaling behavior of SLIP with more compute (longer training) and larger vision models.
We note that 100 epochs of training on YFCC15M corresponds to around 1200 epochs of training on ImageNet-1K.
In Table \ref{tab:scaling} we experimented with holding model size fixed (ViT-B/16) and training for longer as well as training different model sizes for an extended training schedule (100 epochs).
Our results indicate that SLIP scales well with both longer training and larger models.
We show full results simultaneously varying model and compute scaling with SLIP in the appendix.

\begin{table}[h]
\tablestyle{2pt}{1.1}
\setlength{\tabcolsep}{6pt}
  \begin{subtable}[h]{0.45\textwidth}
    \centering
    \begin{tabular}{l c c c c} 
      Model & \#params. & 0-shot & Linear & Finetuned \\
      \shline
      ViT-S	& 22M &	39.5	&	68.3	&	80.7	\\
      ViT-B	& 86M &	45.0	&	73.6	&	83.4	\\
      ViT-L	& 307M &	47.9	&	75.1	&	84.8	\\
    \end{tabular}
    \caption{Comparising ViT model variants of different capacities (ViT-S/B/L). All models are pre-trained for 100 epochs. }
    \vspace{1.0em}
  \end{subtable}
  \hfill
  \begin{subtable}[h]{0.45\textwidth}
      \centering
      \begin{tabular}{l c c c} 
        Epochs & 0-shot & Linear & Finetuned \\
        \shline
        25 &	42.8	&	72.1	&	82.6	\\
        50 & 44.1	&	73.0	&	82.9	\\
        100 & 45.0	&	73.6	&	83.4	\\
      \end{tabular}
    \caption{ViT-B/16 with longer pre-training schedules (25/50/100 epochs).}
  \end{subtable}
  \caption{SLIP pre-training performance (in terms of zero-shot transfer, linear classification, and end-to-end finetuning) can scale well with both model size and number of training epochs.}
  \label{tab:scaling}
\end{table}

\subsection{Additional Benchmarks}

\begin{table*}
\setlength{\tabcolsep}{2pt}
\linespread{1}
\scriptsize
\begin{tabular}{ll cccccccccccccccccccccccccc|c} &&
\rotatebox[origin=lb]{90}{\smash{Food-101}} & \rotatebox[origin=lb]{90}{\smash{CIFAR-10}} & \rotatebox[origin=lb]{90}{\smash{CIFAR-100}} & \rotatebox[origin=lb]{90}{\smash{CUB}} & \rotatebox[origin=lb]{90}{\smash{SUN397}} &
\rotatebox[origin=lb]{90}{\smash{Cars}} & \rotatebox[origin=lb]{90}{\smash{Aircraft}} & \rotatebox[origin=lb]{90}{\smash{DTD}} & \rotatebox[origin=lb]{90}{\smash{Pets}} & \rotatebox[origin=lb]{90}{\smash{Caltech-101}} &
\rotatebox[origin=lb]{90}{\smash{Flowers}} & \rotatebox[origin=lb]{90}{\smash{MNIST}} & \rotatebox[origin=lb]{90}{\smash{FER-2013}} & \rotatebox[origin=lb]{90}{\smash{STL-10}} & \rotatebox[origin=lb]{90}{\smash{EuroSAT}} &
\rotatebox[origin=lb]{90}{\smash{RESISC45}} & \rotatebox[origin=lb]{90}{\smash{GTSRB}} & \rotatebox[origin=lb]{90}{\smash{KITTI}} & \rotatebox[origin=lb]{90}{\smash{Country211}} & \rotatebox[origin=lb]{90}{\smash{PCAM}} &
\rotatebox[origin=lb]{90}{\smash{UCF101}} & \rotatebox[origin=lb]{90}{\smash{Kinetics700}} & \rotatebox[origin=lb]{90}{\smash{CLEVR}} & \rotatebox[origin=lb]{90}{\smash{HatefulMemes}} & \rotatebox[origin=lb]{90}{\smash{SST2}} &
\rotatebox[origin=lb]{90}{\smash{ImageNet}} & \rotatebox[origin=lb]{90}{\smash{Average}} \\
\shline
\multirow{3}{2em}{\rotatebox[origin=c]{90}{ViT-S}}  & CLIP          &	43.4	        &	61.0	        &	29.9	        &	31.1	        &	43.9	        &	3.1	          &	4.7	          &	17.9	        &	25.0	        &	53.3	        &	47.8	        &	9.8	          &	29.1	        &	86.8	        &	22.3	        &	16.1	        &	9.5	          &	34.1	        &	8.7	          &	\textbf{64.8}	&	26.0	        &	18.8	        &	14.7	        &	\textbf{56.1}	&	49.5	        &	32.7	        &	32.3	\\
                                                    & SLIP (25 ep)	&	51.6	        &	\textbf{73.0}	&	35.4	        &	36.3	        &	49.2	        &	4.2	          &	\textbf{6.1}	&	25.7	        &	30.9	        &	62.8	        &	54.3	        &	\textbf{9.9}	&	31.3	        &	91.6	        &	\textbf{22.4}	&	21.9	        &	11.0	        &	\textbf{39.9}	&	\textbf{9.6}  &	50.8	        &	32.8	        &	22.9	        &	14.8	        &	49.6	        &	\textbf{50.1}	&	38.3	        &	35.6	\\
                                                    & SLIP (100 ep)	&	\textbf{53.0}	&	68.4	        &	\textbf{39.3}	&	\textbf{36.5}	&	\textbf{49.8}	&	\textbf{4.6}  &	5.1	          &	\textbf{26.6}	&	\textbf{33.6}	&	\textbf{68.3}	&	\textbf{55.8}	&	2.7	          &	\textbf{37.8}	&	\textbf{91.9}	&	18.2	        &	\textbf{22.2}	&	\textbf{13.8}	&	38.4	        &	8.5	          &	62.8	        &	\textbf{33.3}	&	\textbf{23.5}	&	\textbf{19.2}	&	51.4	        &	49.4	        &	\textbf{39.5}	&	\textbf{36.7}	\\
\hline
\multirow{3}{2em}{\rotatebox[origin=c]{90}{ViT-B}}  & CLIP          &	50.6	        &	66.0	        &	34.5	        &	38.8	        &	51.1	        &	4.0	          &	5.4	          &	21.2	        &	28.5	        &	60.9	        &	53.3	        &	8.4	          &	17.3	        &	90.5	        &	\textbf{30.2}	&	21.5	        &	6.1	          &	35.1	        &	10.5	        &	53.5	        &	28.5	        &	22.1	        &	10.8	        &	52.4	        &	50.7	        &	37.6	        &	34.2	\\
                                                    & SLIP (25 ep)  &	59.5	        &	78.6	        &	45.2	        &	38.7	        &	\textbf{53.4}	&	5.4	          &	5.7	          &	26.1	        &	31.1	        &	71.0	        &	56.6	        &	9.8	          &	19.6	        &	94.4	        &	20.3	        &	\textbf{28.9}	&	\textbf{14.5}	&	34.0	        &	\textbf{11.6}	&	\textbf{55.4}	&	\textbf{37.7}	&	\textbf{26.9}	&	\textbf{17.5}	&	52.8	        &	\textbf{51.1}	&	42.8	        &	38.0	\\
                                                    & SLIP (100 ep)	&	\textbf{63.3}	&	\textbf{79.2}	&	\textbf{50.4}	&	\textbf{44.7}	&	52.0	        &	\textbf{8.1}	&	\textbf{8.4}	&	\textbf{26.2}	&	\textbf{34.7}	&	\textbf{74.0}	&	\textbf{61.3}	&	\textbf{17.1}	&	\textbf{40.8}	&	\textbf{95.4}	&	20.8	        &	27.8	        &	11.7	        &	\textbf{35.2}	&	11.5	        &	52.1	        &	37.1        	&	25.8	        &	13.0	        &	\textbf{55.1}	&	49.9	        &	\textbf{45.0}	&	\textbf{40.0}	\\
\hline
\multirow{3}{2em}{\rotatebox[origin=c]{90}{ViT-L}}  & CLIP          &	59.5	        &	72.9	        &	41.5	        &	\textbf{40.3}	&	53.6	        &	6.9	          &	6.4	          &	20.6	        &	27.9	        &	65.4	        &	55.0	        &	10.3	        &	34.5	        &	94.2	        &	22.7	        &	28.8	        &	5.8	          &	\textbf{41.4}	&	12.6	        &	\textbf{54.9}	&	34.3	        &	24.0	        &	12.9	        &	54.3	        &	\textbf{50.1}	&	40.4	        &	37.4	\\
                                                    & SLIP (25 ep)	&	64.4	        &	\textbf{87.8}	&	\textbf{56.4}	&	39.8	        &	\textbf{58.9}	&	8.6	          &	7.8	          &	26.8	        &	32.0	        &	76.6	        &	59.4	        &	13.2	        &	36.0	        &	\textbf{96.6}	&	27.7	        &	36.5	        &	7.2	          &	28.8	        & \textbf{15.6}	&	54.4	        &	42.6	        &	30.0        	&	14.1	        &	53.4	        &	\textbf{50.1}	&	46.2	        &	41.2	\\
                                                    & SLIP (100 ep)	&	\textbf{69.2}	&	87.5	        &	54.2	        &	39.8	        &	56.0	        &	\textbf{9.0}	&	\textbf{9.5}	&	\textbf{29.9}	&	\textbf{41.6}	&	\textbf{80.9}	&	\textbf{60.2}	&	\textbf{14.9}	&	\textbf{39.6}	&	96.2	        &	\textbf{34.5}	&	\textbf{46.0}	&	\textbf{8.6}	&	30.7	        &	14.2	        &	50.6	        &	\textbf{44.1}	&	\textbf{30.5}	&	\textbf{17.4}	&	\textbf{55.0}	&	49.8	        &	\textbf{47.9}	&	\textbf{43.0}	\\
\hline
\end{tabular}
\vspace{-.5em}
\caption{Zero-shot transfer evaluation with ViT S, B, and L on a variety of classification benchmarks. Best results in \textbf{bold}. SLIP outperforms CLIP on most of the tasks, frequently with a significant margin. With longer pre-training epochs, the performance can be further improved.}
\label{tab:downstream-zeroshot}
\vspace{-.5em}
\end{table*}

While evaluating classification performance on ImageNet gives a broad overview of representation quality, it is also informative to measure performance on a variety of narrowly targeted downstream datasets.
In Table \ref{tab:downstream-zeroshot} we evaluate zero-shot transfer on a battery of downstream image classification tasks compiled by \cite{Radford2021LearningTV}.
These datasets span many different domains including everyday scenes such as traffic signs, specialized domains such as medical and satellite imagery, video frames, rendered text with and without visual context, and more.
We remove Pascal VOC and replace NABirds with CUB-200-2011.
To preprocess the datasets into a unified pipeline we use the extra scripts included in VISSL \cite{goyal2021vissl}.
We catalog chance performance along with short descriptions of the datasets in the appendix.

For a given model size, the relative ranking between methods appears surprisingly inconsistent.
On a few datasets such as Rendered SST2, KITTI depth, and PatchCamelyon (PCAM) it appears at first glance that smaller models and less training improve performance.
However we note that performance on these datasets is only around chance performance, likely because these datasets have little overlap with the semantic distribution of YFCC15M, and thus an unreliable indicator of representation quality.
Performance is stronger on categories are well represented in YFCC15M, such as Food-101, Oxford Pets, Caltech-101, and STL-10.
On these datasets we see that larger models and training for longer with SLIP more generally improve zero-shot transfer accuracy.
We view these results on tasks with more reasonable representation in YFCC15M as more informative of representation quality.

Zero-shot performance on the low-resolution datasets (MNIST, CIFAR-10, CIFAR-100) is also very poor.
On many datasets performance is several multiples of chance performance yet still much lower than what is achievable with a lightweight model trained on a modest amount of application-specific data.
This suggests that language supervision alone is an inefficient way of training models for specific tasks of interest.

We also provide linear classification results on these benchmarks in the appendix.
 
\subsection{Additional Pre-training Datasets}

\begin{table}[h]
\tablestyle{2pt}{1.1}
\begin{tabular}{l l c c c}
Dataset & Method & 0-shot & Linear & Finetuned \\
\shline
CC3M & CLIP & 17.1 & 53.3 & 79.5 \\
& SLIP & 23.0 & 65.4 & 81.4 \\
\hline
CC12M & CLIP & 36.5 & 69.0 & 82.1 \\
  & SLIP & 40.7 & 73.7 & 83.1 \\
\hline
YFCC15M & CLIP & 37.6 & 72.1 & 80.5 \\
& SLIP & 42.8 & 66.5 & 82.6 \\
\end{tabular}
\vspace{-.5em}
\caption{SLIP \vs CLIP ImageNet with ViT-B/16 on CC3M \cite{Sharma2018ConceptualCA}, CC12M \cite{Changpinyo2021Conceptual1P}, two smaller uncurated datasets.
}
\label{tab:datasets}
\vspace{-0.5em}
\end{table}

In addition to YFCC15M, we experiment with two additional image-text datasets: CC12M and CC3M.
In Table \ref{tab:datasets}, we train ViT-B/16 with both SLIP and CLIP on CC12M and CC3M, and compare against our previous numbers on YFCC15M.
SLIP maintains its margin of improvement over CLIP in all ImageNet evaluation settings.
Notably, pre-training SLIP on CC12M instead of YCC15M yields lower zero-shot accuracy but actually results in higher linear and finetuning performance.
CLIP sees an even more surprising boost to finetuning performance of +1.6\%.

Our improved training recipe (see Section \ref{sec:implementation}) largely alleviates overfitting by CLIP on YFCC15M and CC12M, but on the smaller CC3M dataset CLIP overfits quite dramatically.
This may be due to the hypernymization used in CC3M to make the captions more amenable to image captioning.
CLIP reaches its highest zero-shot ImageNet accuracy after just 15 out of 40 epochs of training on CC3M, after which we observe a steady decline in ImageNet accuracy.
In contrast, on CC3M SLIP reaches its highest zero-shot ImageNet performance after 35 epochs.

\subsection{Alternative Self-Supervised Frameworks}

\begin{table}[h]
\tablestyle{2pt}{1.1}
\begin{tabular}{l l c c c}
Method & 0-shot & Linear & Finetuned \\
\shline
SLIP-SimCLR \cite{Chen2020ASF} & 42.8 & 72.1 & 82.6 \\
SLIP-MoCo v3 \cite{Chen2021AnES} & 41.8 & 71.4 & 82.4 \\
SLIP-BYOL \cite{Grill2020BootstrapYO} & 41.3 & 71.1 & 82.2 \\
SLIP-BEiT \cite{Bao2021BEiTBP} & 39.1 & 66.5 & 82.2 \\
\hline
None (CLIP) & 37.6 & 66.5 & 80.5 \\
\end{tabular}
\vspace{-.5em}
\caption{We evaluate ViT-B/16 with several SLIP variants using different self-supervised frameworks.
SLIP works the best with SimCLR among several other self-supervised frameworks, but all variants outperform CLIP.}
\label{tab:frameworks}
\vspace{-0.5em}
\end{table}

As noted in Section \ref{sec:slip_ssl}, SLIP enables the use of many different self-supervision methods.
We ran several experiments on ViT-B/16 with different alternatives to SimCLR, in particular MoCo v3 \cite{Chen2021AnES}, BYOL \cite{Grill2020BootstrapYO}, and BeiT \cite{Bao2021BEiTBP}.
Similar to how we tuned the hyperparameters for SLIP-SimCLR, we largely keep the original self-supervised hyperparameters and add in the CLIP objective and text encoder.
MoCo v3 and BeiT are already designed for ViT, but with BYOL we tuned the learning rate and weight decay while copying the data augmentation and projector/predictor architecture from MoCo v3.
We also lightly tune different scaling parameters for the self-supervised loss.
All models are trained for 25 epochs on YFCC15M.

Our results in Table \ref{tab:frameworks} show that all three alternatives underperform SLIP-SimCLR, despite being individually stronger self-supervised methods.
Most surprising is the result that SLIP-BEiT performs the worst despite BEiT being the strongest self-supervised method tested here.
This may be due to a greater input discrepancy between pre-training and deployment stage. Nonetheless, all these suboptimal variants of SLIP still improve performance over CLIP.

\section{Further Analysis}
\label{sec:analysis}
\paragraph{Why not pre-train with SSL and finetune with CLIP?}
An alternative to SLIP would be to simply initialize the image encoder of CLIP with SSL-trained weights.
We tried training CLIP ViT-B/16 under this setting but found worse performance than training jointly with CLIP and SSL.
Progress after a few training epochs exceeds that of SLIP at the same point in training, but stalls throughout the rest of training (25 epochs).
In \ref{tab:finetune-or-multitask}, we see this approach underperforms SLIP in all three ImageNet evaluation settings.

\paragraph{Is SLIP just CLIP with data augmentation?}
We examine the effects of adding further data augmentation to CLIP and whether this explains the performance improvements seen in SLIP.
The SimCLR augmentation can be separated into two components: color (jitter or grayscale) + blur, and resize crop + flip.
We train CLIP with these two components individually and also with the full SimCLR augmentation.
When training with color + blur, we use the original CLIP cropping strategy from \cite{Radford2021LearningTV} in which we resize the shorter side to 224px then perform a random square crop.
Our results are shown in Table \ref{tab:augmentations}.
While augmentation and resize crop + flip hurt performance, color + blur do improve zero-shot transfer performance by +0.8\% which is still far below the gain by SLIP.

\paragraph{Can we fully decouple self-supervision from language supervision?}
We experimented with a version of SLIP we call SLIP-decoupled in which the self-supervised objective is computed on a disjoint set of 15M images from the YFCC15M images used in the text supervision object.
During training, the images are sampled independently from both sets, effectively decoupling the language-image supervision and self-supervision signals.
In Table \ref{tab:decoupled}, we find that SLIP-decoupled does just as well as SLIP.

\begin{table}[ht]
\tablestyle{2pt}{1.1}
\begin{tabular}{l l l l c c c}
Method & 0-shot & Linear & Finetuned \\
\shline
SimCLR $\rightarrow$ CLIP & 41.1 & 68.2 & 82.3 \\
SLIP-SimCLR & \textbf{42.8} & \textbf{72.1} & \textbf{82.6} \\
\end{tabular}
\vspace{-.5em}
\caption{Finetuning vs. multi-task training.
One alternative to SLIP consists of initializing the image encoder of CLIP with weights trained through self-supervised learning.
With ViT-B/16 trained for 25 epochs, finetuning with CLIP performs noticeably worse across all three ImageNet evaluating settings.
}
\label{tab:finetune-or-multitask}
\vspace{-.5em}
\end{table}

\begin{table}[ht]
\tablestyle{2pt}{1.1}
\setlength{\tabcolsep}{4pt}
\begin{tabular}{l c c c}
Augmentation & 0-shot & Linear & Finetuned \\
\shline
global crop (CLIP) & 37.6 & 66.5 & 80.5 \\
color + blur & 38.4 & 68.5 & 81.5 \\
resize crop + flip & 36.0 & 66.1 & 80.5 \\
color + blur + resize crop + flip & 36.3 & 65.2 & 80.6 \\
\hline
SLIP & \textbf{42.8} & \textbf{72.1} & \textbf{82.6} \\
\end{tabular}
\vspace{-.5em}
\caption{We train CLIP with different data augmentations and compare ImageNet performance to SLIP.
Color + blur slightly improve performance over our improved training recipe using global image crops, but by a much smaller margin than SLIP does.
}
\label{tab:augmentations}
\vspace{-.5em}
\end{table}

\begin{table}[ht]
\tablestyle{2pt}{1.1}
\setlength{\tabcolsep}{4pt}
\begin{tabular}{l c c c}
Method & 0-shot & Linear & Finetuned \\
\shline
SLIP & 42.8 & 72.1 & 82.6 \\
Decoupled SLIP & 42.7 & 72.0 & 82.8 \\
\end{tabular}
\vspace{-.5em}
\caption{Decoupling self-supervision and text-supervision has no effect on performance.
We sampled an additional 15M images disjoint from the YFCC15M images to use only in the self-supervised objective and observe that this performs nearly identically.
}
\label{tab:decoupled}
\vspace{-.5em}
\end{table}

\section{Discussion}
\label{sec:discussion}

Our results on ImageNet and other classification benchmarks show that language supervision and self-supervision are indeed highly synergistic.
As shown in Table \ref{tab:table1}, SLIP improves zero-shot ImageNet performance across model sizes by large margins of +4.8\% to +5.6\%.
Similar gains can be seen in the linear classification setting, with consistent but marginal improvements in the end-to-end finetuning setting.

These trends remain consistent on longer training schedules with the exception of linear probe performance on SLIP ViT-L which actually decreases with more training.
With SLIP ViT-L pre-trained on YFCC15M for 100 epochs, we achieve our strongest result of 47.9\% zero-shot accuracy on ImageNet.
SLIP also shows significant improvements on CC3M and CC12M.
Finally, we also confirm our findings with zero-shot and linear evaluations on additional downstream benchmarks.

\paragraph{Evaluating representation quality.}
Prior work on representation learning has argued against end-to-end finetuning for its sensitivity to optimization hyperparameters \cite{Goyal2019ScalingAB}, and  against linear classification for being too contrived \cite{Zhai2019ALS}.
We note that zero-shot transfer, along with linear classification and end-to-end finetuning, can be viewed as one cohesive paradigm for evaluating representation quality.
Zero-shot transfer represents the strictest setting, where the exemplar vector for each class must be specified through natural language.
Linear classification is a relaxation of zero-shot transfer, in which the class exemplars are optimized on training data.
Finally, end-to-end finetuning represents a further relaxation of linear classification where all model parameters are allowed to adapt to training data.
Representation quality should be assessed by performance across multiple settings, in the same way that an ROC curve offers a more holistic account of model performance than evaluation at a single operating point.

\paragraph{Zero-shot ImageNet monitor.}
SLIP may also serve as a useful framework within which to evaluate new methods for self-supervised learning.
Training loss on the pre-text task is a poor predictor of downstream performance, so a simple external metric like kNN accuracy is important for quickly estimating performance and diagnosing training issues such as overfitting or instability.
However, kNN classification requires encoding and storing every single training image and naive inference requires very expensive matrix multiplications.
The memory bank kNN monitor \cite{Chen2021AnES} alleviates this cost but is not feasible when pre-training on unlabeled datasets such as YFCC100M.
Instead, zero-shot evaluations on ImageNet are virtually as fast as evaluating validation accuracy in the supervised setting.

\paragraph{Ethical considerations.}
SLIP faces all of the same ethical considerations as CLIP, both in terms of the harmful applications it may enable, as well as the potential for amplifying and perpetuating problematic behavior in the real world.
CLIP's ability to leverage noisy and minimally filtered data scraped from the open internet has already spurred researchers to begin collecting data in a more careless manner than previously possible for supervised learning \cite{Birhane2021MultimodalDM}.
A more cautious and responsible approach to selecting training data may alleviate the most egregious model behaviors.

\paragraph{Practical limitations.}
SLIP computes embeddings of image views for both the self-supervised objective and the CLIP objective.
This increases the activation count and memory footprint of the model during the forward pass, which results in slower training (30.5 hours for SLIP vs 22.3 hours for CLIP to train ViT-B/16 on 64 V100 GPUs).
After pre-training, SLIP incurs no additional cost since its vision backbone can be used in the same manner as CLIP or a self-supervised model.

From the downstream results in Table \ref{tab:downstream-zeroshot}, we note that pre-training on uncurated data alone appears to be an inefficient route to recognizing specific visual concepts, especially concepts unlikely to be widely shared on social media or the broader internet.
Even with a massive amount of curated data CLIP's zero-shot performance on many datasets is still far below what can easily be achieved by finetuning a small pre-trained model on a modest amount of labeled data.
This can be easily addressed by simply finetuning CLIP for specific applications or even including more pre-training data from the domain of interest.

\vfill
{\small\bibliographystyle{ieee_fullname}\bibliography{slip}}

\clearpage
\newpage
\appendix

\section{Additional Implementation Details}

\paragraph{Datasets.}
YFCC15M \cite{Radford2018ImprovingLU, Thomee2016YFCC100MTN} contains raw HTML captions and titles which we lightly preprocess before training.
We unescape the HTML then remove HTML tags and urls with simple regex matching.

CC3M \cite{Sharma2018ConceptualCA} is collected from an initial set of 5B candidate images, of which 99.9\% are filtered out according to simple image and text heuristics for quality and content.
Many of these filters are relaxed by CC12M \cite{Changpinyo2021Conceptual1P} in order to collect a bigger and potentially noisier dataset.
CC3M also hypernymizes proper nouns, numbers, and infrequent entities to make the dataset more amenable to training and evaluating image captioning systems, the original design for the dataset.
In contrast, CC12M only replaces person names for privacy.
Our versions of these datasets contain 3.1M and 11.0 M images respectively, due to asset removal.

\paragraph{Pre-training.}
During pre-training we use a cosine learning rate decay schedule with 1 epoch (\app 3500 iterations) of linear warmup when training on YFCC15M.
When pre-training for 100 epochs we use 2 warmup epochs.
On YFCC15M (14.6M images), we train for 25 epochs and on CC12M (11.0M images) we train for 35 epochs.
This amounts to approximately the same number of iterations as 300 epochs on ImageNet-1K~\cite{Russakovsky2015ImageNetLS}.
Due to the smaller size of CC3M (3.1M images), we train for 40 epochs to reduce overfitting.
We trained on up to sixteen 8$\times$ V100-32GB servers, and to fit SLIP ViT-Large/16 in memory we accumulated gradients over two steps.

\paragraph{End-to-end Finetuning.}
We use a similar training recipe for finetuning all models on ImageNet based on the ImageNet finetuning recipe from BeiT~\cite{Bao2021BEiTBP} using AdamW and a batch size of 1024 with learning rate of 4e-3 and weight decay of 0.05, along with various data augmentations and regularization methods.
As we increase model size we also increase regularization.
For ViT-S we set drop path to 0 and layer decay to 0.65, for ViT-B we set drop path to 0.1 and layer decay to 0.65, and for ViT-L we set drop path to 0. 

\section{Full Scaling Results}



We include the full results of our scaling experiments in Table~\ref{tab:full-scaling}, in which we simultaneously increase model size and training epochs.
As measured by ImageNet classification accuracy under the three settings (zero-shot transfer, linear classification, and end-to-end finetuning), both large models and longer training generally improve performance.

The exception to this trend is the linear classification performance of SLIP ViT-L/16, which degrades slightly with longer training.
This behavior also persists across the various other downstream benchmarks, where SLIP ViT-L/16 does worse on average when trained for 100 epochs than when trained for 25 epochs.
We note that both the zero-shot transfer and end-to-end finetuning performance of SLIP ViT-L/16 improve with longer training, contrary to the behavior seen with linear classification.
Thus we cannot declare this behavior to be a case of simple overfitting, as the representations are still improved for the other evaluation settings.

\section{Additional Linear Classification Benchmarks}
In Table~\ref{tab:downstream-linear} we show linear classification results on all 26 downstream datasets (including ImageNet).
With ViT-B and ViT-S, SLIP pre-training for 100 epochs does best.
As with ImageNet, SLIP ViT-L also does worse on average when trained for 100 epochs than when trained for 25 epochs.
The dataset average is 0.5 points lower for the 100 epoch model.

As expected, linear classification accuracy is much higher than zero-shot transfer accuracy (shown in Table \ref{tab:downstream-zeroshot}).
However, the gap between zero-shot and linear performance varies between datasets.
On datasets which are straightforward vision tasks but poorly represented among the YFCC100M imagery, such as Patch Camelyon, MNIST, KITTI distance, and GTSRB, linear classification massively improves accuracy, often from a baseline of around chance performance.
On datasets which share more overlap with YFCC100M, such as Food-101, Caltech-101, and Caltech-UCSD Birds 2011, we see significant improvements as well.

However, with HatefulMemes and Rendered SST2, two datasets which require OCR capabilities, the linear classification performance of all models is still around chance.
These results suggest, perhaps unsurprisingly, that zero-shot transfer results are much more dependent on what visual and semantic concepts were seen during training than linear classification, since they do not enjoy the benefit of further training examples.
We also note that relative rankings within each model size are also quite unstable where the best results alternate between the 25 and 100 epoch models.
This is very similar to what we see in the zero-shot transfer evaluations, as discussed in Section \ref{sec:evaluations}.

\begin{table}[hbt]
\tablestyle{4pt}{1.1}
\begin{tabular}{l | ccc | ccc | ccc} 
& \multicolumn{3}{c}{0-shot} & \multicolumn{3}{c}{Linear} & \multicolumn{3}{c}{Finetuned} \\
Model & 25 & 50 & 100 & 25 & 50 & 100 & 25 & 50 & 100 \\
\shline
ViT-S/16	&	38.3	&	39.3	&	39.5	&	66.4	&	67.6	&	68.3	&	80.3	&	80.7	&	80.7	\\
ViT-B/16	&	42.8	&	44.1	&	45.0	&	72.1	&	73.0	&	73.6	&	82.6	&	82.9	&	83.4	\\
ViT-L/16	&	46.2	&	47.4	&	47.9	&	76.0	&	75.8	&	75.1	&	84.2	&	84.7	&	84.8	\\
\end{tabular}
\caption{Full scaling experiment results.
SLIP scales well to larger models and longer training as measured by zero-shot transfer, linear classification, and end-to-end finetuning, with the exception of linear classification performance using ViT-L.
}
\label{tab:full-scaling}
\end{table}

\begin{table*}[ht]
\setlength{\tabcolsep}{2pt}
\linespread{1}
\scriptsize
\begin{tabular}{ll cccccccccccccccccccccccccc|c} &&
\rotatebox[origin=lb]{90}{\smash{Food-101}} & \rotatebox[origin=lb]{90}{\smash{CIFAR-10}} & \rotatebox[origin=lb]{90}{\smash{CIFAR-100}} & \rotatebox[origin=lb]{90}{\smash{CUB}} & \rotatebox[origin=lb]{90}{\smash{SUN397}} &
\rotatebox[origin=lb]{90}{\smash{Cars}} & \rotatebox[origin=lb]{90}{\smash{Aircraft}} & \rotatebox[origin=lb]{90}{\smash{DTD}} & \rotatebox[origin=lb]{90}{\smash{Pets}} & \rotatebox[origin=lb]{90}{\smash{Caltech-101}} &
\rotatebox[origin=lb]{90}{\smash{Flowers}} & \rotatebox[origin=lb]{90}{\smash{MNIST}} & \rotatebox[origin=lb]{90}{\smash{FER-2013}} & \rotatebox[origin=lb]{90}{\smash{STL-10}} & \rotatebox[origin=lb]{90}{\smash{EuroSAT}} &
\rotatebox[origin=lb]{90}{\smash{RESISC45}} & \rotatebox[origin=lb]{90}{\smash{GTSRB}} & \rotatebox[origin=lb]{90}{\smash{KITTI}} & \rotatebox[origin=lb]{90}{\smash{Country211}} & \rotatebox[origin=lb]{90}{\smash{PCAM}} &
\rotatebox[origin=lb]{90}{\smash{UCF101}} & \rotatebox[origin=lb]{90}{\smash{Kinetics700}} & \rotatebox[origin=lb]{90}{\smash{CLEVR}} & \rotatebox[origin=lb]{90}{\smash{HatefulMemes}} & \rotatebox[origin=lb]{90}{\smash{SST2}} &
\rotatebox[origin=lb]{90}{\smash{ImageNet}} & \rotatebox[origin=lb]{90}{\smash{Average}} \\
\shline
\multirow{3}{2em}{\rotatebox[origin=c]{90}{ViT-S}} & CLIP & 71.1          & 82.5          & 63.2          & 66.5          & 70.7          & 28.2          & 26.0          & 61.6          & 64.1          & 82.5          & 91.8          & 89.6          & 45.2          & 92.1          & 92.4          & 83.6          & 65.7          & 63.6          & 22.0          & 80.4          & 67.7          & 31.8          & 44.3          & 53.0          & 51.0          & 59.3          & 63.5          \\
& SLIP (25 ep) & 77.4          & 80.7          & 63.5          & \textbf{67.0} & \textbf{74.2} & 39.3          & \textbf{33.8} & 70.7          & 69.5          & 87.2          & 94.5          & \textbf{91.3} & \textbf{52.7} & 95.7          & 94.3          & \textbf{90.5} & \textbf{68.2} & 65.8          & 22.4          & 81.9          & 76.9          & 39.2          & \textbf{51.6} & \textbf{55.6} & \textbf{54.3} & 66.4          & 67.9          \\
& SLIP (100 ep) & \textbf{78.7} & \textbf{84.1} & \textbf{66.3} & 66.0          & 73.9          & \textbf{40.6} & 32.6          & \textbf{71.6} & \textbf{72.0} & \textbf{87.9} & \textbf{95.3} & 90.7          & 50.4          & \textbf{96.4} & \textbf{95.2} & 89.0          & \textbf{68.2} & \textbf{66.8} & \textbf{23.3} & \textbf{82.9} & \textbf{77.7} & \textbf{40.5} & 50.5          & 53.4          & 53.4          & \textbf{68.3} & \textbf{68.3} \\
\hline
\multirow{3}{2em}{\rotatebox[origin=c]{90}{ViT-B}} & CLIP & 77.6          & 86.2          & 70.7          & 70.9          & 73.7          & 41.8          & 32.0          & 66.0          & 70.5          & 85.3          & 94.2          & 93.8          & 49.7          & 94.9          & 94.5          & 88.3          & 72.5          & 65.8          & 24.9          & 82.9          & 72.6          & 36.4          & 48.7          & 53.4          & 54.8          & 66.5          & 68.0          \\
& SLIP (25 ep) & 83.0          & 87.7          & \textbf{71.6} & 70.9          & 76.3          & 47.4          & 36.8          & 73.9          & 74.4          & 89.7          & \textbf{96.4} & \textbf{94.5} & 53.5          & 97.4          & \textbf{95.9} & \textbf{92.8} & 75.5          & \textbf{68.6} & 25.1          & \textbf{84.4} & 80.4          & 43.7          & \textbf{54.2} & \textbf{57.8} & 55.0          & 72.1          & 71.5          \\
& SLIP (100 ep) & \textbf{83.1} & \textbf{88.9} & 71.5          & \textbf{72.0} & \textbf{76.4} & \textbf{49.0} & \textbf{37.0} & \textbf{75.9} & \textbf{75.8} & \textbf{90.7} & 96.2          & \textbf{94.5} & \textbf{54.4} & \textbf{98.2} & 95.5          & 92.0          & \textbf{75.9} & 67.9          & \textbf{25.6} & 83.0          & \textbf{82.0} & \textbf{44.4} & 53.9          & 55.2          & \textbf{56.1} & \textbf{73.6} & \textbf{71.9} \\
\hline
\multirow{3}{2em}{\rotatebox[origin=c]{90}{ViT-L}} & CLIP & 81.8          & 91.2          & 75.1          & 75.1          & 75.4          & 46.9          & 34.7          & 66.2          & 73.0          & 86.4          & 95.4          & 95.7          & 54.7          & 96.5          & 95.1          & 90.6          & 76.4          & 68.7          & 27.2          & 83.6          & 75.9          & 39.9          & 51.9          & 57.0          & 53.8          & 70.5          & 70.7          \\
& SLIP (25 ep) & \textbf{86.5} & \textbf{92.9} & \textbf{77.2} & \textbf{76.6} & \textbf{78.0} & \textbf{52.0} & \textbf{38.8} & 75.2          & 79.4          & 90.7          & \textbf{97.7} & \textbf{95.8} & 56.8          & \textbf{98.7} & 96.6          & \textbf{93.3} & \textbf{79.9} & 68.9          & \textbf{29.1} & 84.8          & 83.0          & \textbf{47.7} & \textbf{57.5} & 52.6          & 54.5          & \textbf{76.0} & \textbf{73.9} \\
& SLIP (100 ep) & 84.1          & 91.0          & 74.5          & 72.5          & 76.7          & 51.1          & 38.4          & \textbf{77.0} & \textbf{79.8} & \textbf{91.4} & 97.2          & \textbf{95.8} & \textbf{60.2} & \textbf{98.7} & \textbf{97.0} & 93.2          & 75.6          & \textbf{70.6} & 28.2          & \textbf{85.2} & \textbf{83.1} & 45.8          & 55.0          & \textbf{54.0} & \textbf{56.2} & 75.1          & 73.4         \\
\hline
\end{tabular}
\caption{Linear classification evaluation with ViT S, B, and L on a variety of classification benchmarks. Best results in \textbf{bold}. SLIP outperforms CLIP on most of the tasks, frequently with a significant margin.}
\label{tab:downstream-linear}
\end{table*}

\begin{table*}[h]
\tablestyle{2pt}{1.1}
\begin{tabular}{l l c c c}
Dataset & Metric & Chance performance & Description \\
\shline
Food-101 \cite{bossard14} & acc & 1.0 & 101 categories of food dishes \\
CIFAR-10 \cite{Krizhevsky2009LearningML} & acc & 10.0 & 10 categories of animals and vehicles \\
CIFAR-100 \cite{Krizhevsky2009LearningML} & acc & 1.0 & 100 categories of animals, vehicles, plants, objects, scenes, people \\
CUB-200-2011 \cite{WahCUB_200_2011} & acc & 0.8 & 200 species of mostly North American birds\\
SUN397 \cite{Xiao2010SUNDL} & acc & 2.2 & 397 categories of various indoor and outdoor scenes \\
Stanford Cars \cite{KrauseStarkDengFei-Fei_3DRR2013} & acc & 0.8 & 196 categories of cars (make, model, and year) \\
FGVC Aircraft \cite{maji13fine-grained} & mean per class & 1.0 & 102 categories of aircraft (manufacturer, family, and variant) \\
Describable Textures \cite{cimpoi14describing} & acc & 2.1 & 47 categories of texture patches \\
Oxford Pets \cite{Parkhi2012CatsAD} & mean per class & 2.7 & 37 breeds of cats and dogs \\
Caltech-101 \cite{FeiFei2006OneshotLO} & mean per class & 5.2 & 101 categories of objects \\
Oxford Flowers \cite{Nilsback2008AutomatedFC} & mean per class & 1.5 & 102 species of common UK flowers \\
MNIST \cite{LeCun1998GradientbasedLA} & acc & 10.0 & 10 categories of handwritten digits\\
FER-2013 \cite{Goodfellow2013ChallengesIR} & acc & 24.7 & 7 categories of human facial emotions\\
STL-10 \cite{Coates2011AnAO} & acc & 11.4 & 10 categories of animals and vehicles \\
EuroSat \cite{Helber2019EuroSATAN} & acc & 10.0 & 10 categories of land from satellite imagery \\
RESISC45 \cite{Cheng2017RemoteSI} & acc & 2.2 & 45 categories of land from satellite imagery and aerial photography\\
GTSRB \cite{Stallkamp2011TheGT} & acc & 5.9 & 43 categories of German traffic signs \\
KITTI Distance \cite{Geiger2012AreWR} & acc & 31.0 & 4 categories of traffic scenes with nearby cars in varying positions \\
Country211 \cite{Radford2021LearningTV, Thomee2016YFCC100MTN} & acc & 0.5 & 211 countries represented by geo-tagged images \\
Patch Camelyon \cite{Veeling2018-qh, EhteshamiBejnordi2017DiagnosticAO} & acc & 50.0 & 2 classes of metastatic or benign lymph node slide patches \\
UCF101 Frames \cite{Soomro2012UCF101AD} & acc & 1.3 & 101 categories of human actions using the middle frame of each clip \\
Kinetics 700 Frames \cite{Carreira2019ASN} & mean(acc1, acc5) & 0.4 & 700 categories of human actions using the middle frame of each clip \\
Clevr Counts \cite{Johnson2017CLEVRAD} & acc & 12.9 & 8 categories of rendered scenes with varying numbers of objects \\
Hateful Memes \cite{Kiela2020TheHM} & ROC AUC & 50.0 & 2 categories of hateful or not hateful image macros \\
Rendered SST2 \cite{Radford2021LearningTV, Socher2013RecursiveDM} & acc & 50.1 & 2 classes of positive or negative movie reviews rendered as text \\
ImageNet \cite{Russakovsky2015ImageNetLS} & acc & 0.1 & 1000 categories of objects\\
\end{tabular}
\caption{Info sheet for classification datasets. Chance performance is computed by assuming random predictions of the labels in proportion to their frequency in the test set.}
\label{tab:infosheet}
\end{table*}

\end{document}